# Efficient Large-scale Approximate Nearest Neighbor Search on the GPU

Patrick Wieschollek[1,4]   Oliver Wang[2]   Alexander Sorkine-Hornung[3]
Hendrik P.A. Lensch[1]

[1] University of Tübingen   [2] Adobe Systems Inc.
[3] Disney Research
[4] Max Planck Institute for Intelligent Systems, Tübingen

## Abstract

*We present a new approach for efficient approximate nearest neighbor (ANN) search in high dimensional spaces, extending the idea of Product Quantization. We propose a two level product and vector quantization tree that reduces the number of vector comparisons required during tree traversal. Our approach also includes a novel highly parallelizable re-ranking method for candidate vectors by efficiently reusing already computed intermediate values. Due to its small memory footprint during traversal the method lends itself to an efficient, parallel GPU implementation. This Product Quantization Tree (PQT) approach significantly outperforms recent state of the art methods for high dimensional nearest neighbor queries on standard reference datasets. Ours is the first work that demonstrates GPU performance superior to CPU performance on high dimensional, large scale ANN problems in time-critical real-world applications, like loop-closing in videos.*

## 1. Introduction

Finding the nearest neighbors (NN) of a query vector in a high dimensional space is a fundamental task in computer vision. For a given query vector $y \in \mathbb{R}^D$, the nearest neighbor problem consists of finding an element $N(y) \in \mathcal{X}$ from a predefined fixed set $\mathcal{X} \subset \mathbb{R}^D$, which minimizes a distance metric, most commonly the Euclidean distance $d(\cdot) := \|\cdot\|^2$. This NN-problem can be written as

$$N(y) = \arg\min_{x \in \mathcal{X}} d(y, x). \quad (1)$$

In many computer vision tasks these query vectors represent visual descriptors, meaning both $D$ as well as $\mathcal{X}$ are often large, leading to NN searches being a significant computational bottleneck in many applications. This is due to the necessity of computing exact distances in a high dimensional space between many pairs of vectors, a problem exacerbated by the phenomenon known as the *curse of dimensionality*. More precisely, consider a $D$-dimensional hyper unit-cube enclosing $\mathcal{X}$. To explore a $\nu$ fraction of the volume, we need to visit a $\nu^{1/D}$ percent of each hyper-cube edge. This means that to explore 10% of a set of SIFT-vectors ($D = 128$) in a hypercube, one has to search an interval covering $\approx 98\%$ of the possible values per coordinate.

Due to this computational complexity, most applications rely on *approximate* nearest-neighbor (ANN) search techniques, which try to find the nearest neighbor with a high probability. There exists many CPU-approaches for computing ANN in the literature, the most common of which are *KD-trees* [6], which hierarchically subdivide the vector space. While these methods are widely used in graphics and vision, it has been shown that KD-trees are no more efficient than brute force searches when $D$ is large [9]. The FLANN software package [16] proposes randomized KD-forests and K-Means trees, which prune the overall search space by identifying small regions around the query vectors, yielding better performance with higher dimensional vectors.

Another family of approaches are based on *Locality Sensitive Hashing* (LSH) [5]. These methods hash database vectors with a number of random projections, and perform nearest distance checks only on vectors that are hashed to the same bin. The speed and accuracy of such methods depends on the hashing function used. Andoni and Indyk [1] describe a family of hashing functions which are near-optimal. These ideas have since been extended into the image domain for patch-based nearest neighbor computation [13]. While these methods work well, they have not yet achieved the same performance as space partitioning methods [16].

While leveraging GPU parallelism seems obvious, in practice accelerating ANN search techniques using GPU parallelism is notoriously difficult, largely due to the memory restrictions of GPUs when compared to the amount of RAM available to CPUs. As a result, existing GPU-based methods often implement brute force approaches, are limited to small datasets of up to 225 candidate neighbors [18] or can handle only 3-dimensional vectors [16], making these approaches unsuited for many vision problems. Our method is designed to be highly parallel, and can be easily implemented on a GPU for significant improvements in query time, while even a CPU version is competitive to previous methods.

In general, ANN searches are composed of an *offline* phase, containing all query-*independent* computations like building an indexing structure of the dataset; and an *online* phase which comprises all query-*dependent* computations. However, reported results from previous state-of-the-art methods [11, 9, 3, 2] excluded expensive query-*dependent* pre-computations from the timing of their online-phase. This does not give a reasonable expectation of running time, as in nearly all applications, the query vector $y$ is *unknown* until the query request, and cannot be precomputed in an offline phase. Therefore, we include *all* query-dependent computation steps in our timing results, in order to give a better indication as to running times achievable for real applications.

Vector Quantization (VQ) [14] is a simple method that *clusters* the search space into a number of bins based on the distance to the cluster centroid. If a query vector is quantized to a bin, all other vectors in that bin are likely to be good candidates for being the nearest neighbor. Unfortunately, if a query lies at the edge of a bin, one has to consider all neighboring bins as well, and the number of neighbors to each Voronoi cell increases *exponentially* w.r.t to the dimension $D$ of the space.

The concept of *Product Quantization* (PQ) was introduced in [12] and made popular in the computer vision community by Jegou et al. [9]. Several state-of-the-art ANN approaches extend these ideas, such as locally optimized product quantization[11] and the inverted multi-index [4]. These methods currently provide the most efficient techniques for ANN search for high dimensional data, in terms of speed, accuracy, and memory requirements. In general, PQ based approaches consist of the following three main steps: (1) a robust *proposal mechanism* is used to identify a list of nearest neighbor candidates in the database (similar to the vectors from a bin in the VQ example), (2) a *re-ranking* step then sorts these candidates according to their ascending approximate distances to the query vector. Finally, the approximated k-nearest vectors after re-ranking are further sorted using (3) an *exact distance calculation*.

We present an extension to the family of PQ methods called the Product Quantization Tree (PQT). The main contributions of our approach are: a two level product and quantization tree that requires significantly fewer exact distance tests than prior work; a relaxation of the Dijkstra Algorithm for an effective bin traversal order; a fast, re-ranking step to approximate exact distances between query and database points in constant time $\mathcal{O}(P)$, where a query vector is split into $P$ parts; and a highly optimized GPU based open-source implementation.

In this work, we compare our method using a common benchmark, BigANN [9], which consists of 1 billion 128-dimensional SIFT-vectors and 10000 query vectors. This dataset is challenging due to the infeasibility of an exhaustive search, as well as the sheer size of the data (just storing the database vectors requires 132 GB of memory). At comparable approximation quality the GPU implementation achieves significant speed-up over prior work. We provide source code of our approach to encourage the development of new applications that require high-performance ANN queries.

## 2. Background

Our approach builds on PQ [9], which we describe here, followed by a description of the most related work to ours. Let $\mathcal{X} = \{x_1, \ldots, x_n\}$ be a finite set of database vectors $x_i \in \mathbb{R}^D$. Without loss of generality we consider the Euclidean space $(X, d)$, however our approach can be used with any arbitrary metric $d$.

### 2.1. Vector and Product Quantization

In vector quantization (VQ), each vector $x \in \mathbb{R}^D$ is encoded by a codebook $\mathcal{C} = \{c_1, \ldots, c_k\}$ of $k$ centroids with the mapping: $c \colon \mathcal{X} \to \mathcal{C}, \quad x \mapsto c(x) := \arg\min_{c \in \mathcal{C}} d(x, c)$, In other words, each vector is represented by its closest centroid in the codebook. The set $C_k = \{x \in \mathbb{R}^D | c(x) = c_k\}$ is called the *cluster* or *bin* for centroid $c_k$. This quantization of vectors introduces an approximation error, but allows for quick retrieval of a similar set of vectors $C_k$, i.e., all those that are quantized to the same bin as the query. Classical Lloyd iterations [15] can be used on a subset of the original data to efficiently find a good codebook $\mathcal{C}$.

In PQ, the high dimensional vector space is transformed into a product space, whose subspaces are then quantized using VQ. Under the assumption that $D = P \cdot m$ for some $P, m \in \mathbb{N}$ we can write $x \in \mathbb{R}^{P \cdot m}$ as the concatenation of $P$ vector *parts* $x = ([x]_1, [x]_2, \ldots, [x]_P)^T$ with $[x]_i \in \mathbb{R}^m$. This allows for exponentially large codebook generation by encoding $x \in \mathbb{R}^D$ into a Cartesian product of subspaces $\mathcal{C} = \mathcal{C}_1 \times \mathcal{C}_2 \times \cdots \times \mathcal{C}_P$, with $k^P$ bins, while only requiring space for $k \cdot P$ centroids (see Figure 1d) . Increasing the number of bins enables a much finer granularity for the query process, and so the vectors in each single bin have a significant higher correlation. The canonical projection is a mapping of each part $[x]_p$ independently

$$c_p \colon \mathcal{X} \to \mathcal{C}_p, \quad x \mapsto c_p(x) := \arg\min_{c \in \mathcal{C}_p} d([x]_p, c), \quad (2)$$

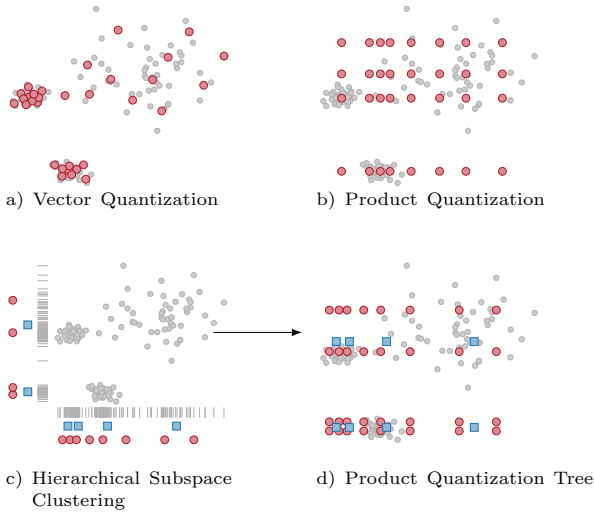

a) Vector Quantization  
b) Product Quantization  
c) Hierarchical Subspace Clustering  
d) Product Quantization Tree

Figure 1: Three different quantization schemes with $k = 32$ clusters. Vector Quantization (a) represents vectors by their closest centroids. Product Quantization performs the clustering in subspaces (here axes) (b). A tree structure can be used to build a hierarchy of clusters on each axis (c). Our method use the hierarchy of two quantization levels, first using PQ with a low number of centroids, and then a second-layer of PQ within these bins (d). Points drawn as ▪ are PQ centroids, and each corresponding cluster is split again into finer 4 clusters (2 on each axis) with centroids illustrated as ●.

to its nearest part-centroid. The nearest centroid $c(x) \in \mathcal{C}$ for $x \in \mathbb{R}^D$ is the concatenation of the sub-centroids

$$c(x) = (c_1(x), c_2(x), \ldots, c_p(x))^T. \quad (3)$$

Finding a good quantizer $c(\cdot)$ for $\mathcal{X}$ is distributed into finding $P$ codebooks $\mathcal{C}_1, \mathcal{C}_1, \ldots, \mathcal{C}_P$ independently, which can also be done using Lloyd iterations. Note that when setting $P = 1$, Product Quantization becomes Vector Quantization.

While it is indeed easy to produce exponentially many (in terms of $P$) clusters using PQ, most will be empty because of the dataset distribution (see supplementary material). However, if we assume that the set of query vectors has a similar distribution as the set of database vectors, we can expect that most queries will also correspond to non-empty clusters. Nonetheless, we still must be able to deal with clusters of highly diverse cardinality as illustrated in Figure 1.

For a better quantization, the authors of [7] proposed to augment PQ using the mapping

$$c \colon \mathcal{X} \to \mathcal{C}, \quad x \mapsto c(x) := \arg\min_{c \in \mathcal{C}} d(Rx, c), \quad (4)$$

where $R \in \mathbb{R}^{D \times D}$ is a rotation matrix. However, this requires a high dimensional matrix multiplication for rotating the query for each visited cluster, which is expensive even for a GPU implementation. We note that the authors of [11] pre-compute all possible projections $Ry_i$ of the query vectors in the offline phase, but this approach is only practical when query vectors are known beforehand.

### 2.2. Inverted file system

For a query vector $y \in \mathbb{R}^D$, the approach of [9] proposes an inverted index system with an asymmetric distance computation. This consists of an initial VQ step that acts as a coarse quantizer with a codebook of $k$ centroids to extract a set of candidates vectors ($k = 8192$ clusters are used [9]). The number of candidates is empirically set to 0.05% of the database size to achieve a recall $\geq 0.9$ by visiting 64 clusters. This corresponds to 524288 candidate vectors for each query in the BigANN database. This approach requires $k$ exact $D$-dimensional distance calculations for each query vector $y$ to identify reasonable clusters. The centroids are then sorted based on distances, and the $w$-best clusters are chosen, giving a list of database vectors $\mathcal{L}_c \subset \mathcal{X}$ which have a high chance of containing the nearest neighbor.

These vectors are again sorted in a *re-ranking* step based on PQ of the expensive residual-computation $r_w = y - c_w$ to the identified cluster $c_w$, which is precomputed in [9], [11] and stored in a distance lookup-table. Again, this pre-computation is only possible when query vectors are known *in advance*.

The distance between the query vector $y \in \mathbb{R}^D$ and each nearest neighbor candidate $x \in \mathcal{L}_c$ can be approximated by quantizing the residual using a second PQ codebook with $k_2$ words. Re-ordering the list $\mathcal{L}_c \to \mathcal{L}_s$ and considering the first few vectors $\mathcal{L}'_s \subset \mathcal{L}_s$, an exhaustive search in $\mathcal{L}'_s$ becomes feasible.

The lookup and re-ranking steps when visiting $w$ clusters requires $k_1 + w \cdot k_2$ exact distance calculations during query time. With typical values of $k_1 = 8192, w = 64, k_2 = 256$, this implies 24576 distance calculations.

Our hierarchical approach reduces the total number of exact distance calculations to less than 200. On a modern NVIDIA Titan X computing 16k exact distances is 62 times slower (0.13 ms) than 256 exact distance computations (0.0021 ms). Hence, the number of exact vector comparison would become a bottleneck. A *complete* query in our algorithm only takes about 0.02 ms due to an efficient index structure and the parallel nature of our approach.

### 2.3. Inverted multi-index (IMI)

The inverted multi-index [4] exploits PQ rather than VQ for an indexing structure over all database vectors, which reduces the number of centroid-distance calculation for cluster proposals or vise-versa increases the number bins: Given part distances to $k$ codebook vectors, for each part $[y]_p$ of

the query vector this approach sorts the corresponding $k$ centroids w.r.t. to the ascending distances

$$
\begin{aligned}
{[y]}_1 &\to \{i_{10}, \quad i_{11}, \quad i_{12}, \quad \ldots, \quad i_{1k-1}\} &= \mathcal{I}_1, \\
{[y]}_2 &\to \{i_{20}, \quad i_{21}, \quad i_{22}, \quad \ldots, \quad i_{2k-1}\} &= \mathcal{I}_2, \\
&\vdots \\
{[y]}_P &\to \{i_{P0}, \quad i_{P1}, \quad i_{P2}, \quad \ldots, \quad i_{Pk-1}\} &= \mathcal{I}_P,
\end{aligned}
$$

where $i_{23}$ is the ID of the 3rd nearest cluster for part 2. The combined cluster IDs of all parts encoded a bin ID via a multi-index

$$i \in \mathcal{I}_1 \times \mathcal{I}_2 \times \cdots \times \mathcal{I}_P. \qquad (5)$$

For NN-search, starting with bin $B_i, i = (i_{10}, i_{20}, \ldots, i_{P0})$ a heuristic is needed to traverse all bins $B_i$ in the vicinity.

The authors of [4] make use of a priority queue to dynamically select the next closest not yet visited bin until sufficiently many bins/vectors are proposed. All vectors stored in each visited bin $B_i$ are then examined in an exhaustive search using PQ-based re-ranking of the residual to each bin centroid.

Both methods [9, 4] achieve state-of-the-art precision but have efficiency issues when making a query. VQ-based indexing requires a very large number of full dimensionality codebook vector comparisons, and even for PQ-based indexing the number is still large. PQ-based indexing is hindered by a slow enumeration of the next best bin [4]. Additionally, both methods require quantizing the residual within each bin for re-ranking; this is accelerated by pre-computing the residual quantization for each part in [4], however with unknown query vectors, this optimization can not be made.

We address these issues by introducing a Product Quantization Tree (Sec. 3). Our approach presents an efficient heuristic for proposing bins (Sec. 3.2), as well as a novel re-ranking method based on projections to quantized lines for re-ranking (Sec. 3.3). Our re-ranking step is especially efficient as it can simply reuse distance calculations computed during the tree traversal. Finally, we demonstrate that our approach can be efficiently implemented on a GPU using CUDA (Sec. 4).

## 3. Product Quantization Tree

The Product Quantization Tree (PQT) is built upon a combination of the inverted multi-index and hierarchical PQ. The main idea is that product quantization is performed using a hierarchical VQ-tree [16] for each part rather than a flat codebook. The tree structure on the centroids speeds up the query (*online*), sorting into the database (*offline*), and indexes considerably more bins in contrast to the inverted multi-index. Additionally, it is designed to enable the reuse of computed values for fast re-ranking.

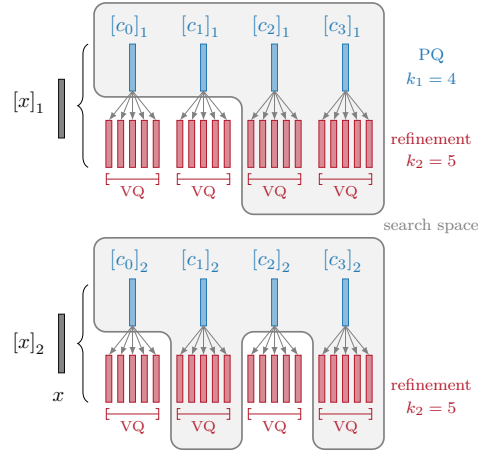

Figure 2: Both parts of $([x]_1, [x]_2) \in \mathbb{R}^D$ are quantized by a VQ tree with $k_1 = 4$ clusters in the first and $k_2 = 5$ finer clusters in the second level. During traversal, only the best $w$ closest clusters of the first level are refined. The example search space by extending $w = 2$ clusters is illustrated as the gray area.

### 3.1. Tree structure - offline phase

Sorting each database vector $x \in \mathcal{X}$ into a bin $B_\ell$ gives disjoint sets, $\mathcal{X} = \bigcup_{k=1}^{K} B_k$. We describe how to effectively map a vector into a bin, $m \colon \mathcal{X} \to \mathcal{I}_1 \times \mathcal{I}_2 \times \cdots \times \mathcal{I}_P$.

The indexing structure is a tree which consists of two levels of quantizers. The first level is a traditional $P$-parts product quantizer with $k_1$ centroids for each part. Each resulting part cluster is then *independently* refined by one additional vector quantizer with $k_2$ centroids as illustrated in Figure 2. The bins are addressed by any combination of the per-part child node centroids. This gives $K = (k_1 \cdot k_2)^P$ bins in total.

**Training the codebook.** Constructing the VQ-trees is done independently for each part, first by constructing a VQ codebook (level 1) using Lloyd iteration in the fashion of the Linde-Buzo-Gray algorithm [14] and then quantizing all sub-vectors (level 2) assigned to a first level cluster.

While the inverted multi-index approach [4] also uses two levels of product quantization, the second is exclusively used for re-ranking. As opposed to this, we use two levels for *indexing*. While additional tree levels are possible, we empirically found this configuration to be optimal in terms of balancing the number of bins to check with the reduction of candidate vectors.

### 3.2. Query - online phase

A query now consist of four steps: tree traversal, bin proposal, vector proposal, and re-ranking.

The tree traversal is carried out as described above producing an ordered list of $(i, d)_p^2$ for the best subset of level 2

clusters.

**Tree Traversal.** The tree reduces the number of exact distance computations required during traversal by pruning. After comparing to all $k_1$ first-level codebook vectors, the distances are sorted, and only the $w$ best clusters are refined for further distance calculations for the level 2 codebook. Let $y \in \mathbb{R}^D$ be the query vector, distances $d([y]_p, [c^1]_p)$ to the $k_1$ first-level . in the first level are computed separately for each part. This step returns a set of IDs and distance pairs

$$\{(i, d')_p^1 \mid d' = d([y]_p, [c_i^1]_p)\} \tag{6}$$

for each part $p$ and each level 1 centroid.

From these possible per-part clusters, we only use the closest $w$ centroids for further processing, i.e. computing the distances $d([y]_p, [c^2]_p)$ only to those level 2 centroids whose corresponding $c^1$ are in the best set. The level 2 distances are ordered to find the best cluster indices for each part. Finally, combining the best indices of the individual parts identifies the best bin as in Equation 5.

A typical configuration might consist of four parts ($P = 4, k_1 = 16, k_2 = 16, w = 4$), amounting to only $16 + 4 \cdot 16 = 80$ full vector distance calculations to address $(16 \cdot 16)^4 \approx 4$ trillion bins. For practical purposes we applied modulo-hashing by using unsigned integers representing the index.

**Bin Proposal Heuristic.** Given the best bin as determined by the index $i = (i_{11}, i_{21}, \ldots, i_{P1})$, one has to find a sequence of neighbor bins to check such that a sufficient number of vectors for re-ranking is generated. The priority queue used in Babenko and Lemptsky [4] would yield the optimal sequence but it requires a resorting operation for each proposed bin, which is expensive and is sequential by nature.

Instead, we propose to choose a fixed traversal heuristic. The most simple order would be to compute all id-vectors $v \in \{1, r\}^P$ and sort them according their distance from the origin $\|v\|_2$. However, this returns an isotropic bin traversal heuristic as depicted in Figure 3 (blue line) compared to the optimal sequence from [4] (green line) and our proposed anisotropic traversal heuristic (red line). The anisotropic version with flexible slope produced a better approximation of the Dijkstra ordering. Hereby, we pre-compute bin orderings for 10 slopes $1.08^k$ with $k = -5, -4, \ldots, 4$. Each slope describes the progress balance on one part-pair. A slope of 1 would equally handle both parts, while a slope of $1.08^{-5}$ would allow more bin combinations with higher ids in the second part (see red line in Figure 3).

### 3.3. Re-ranking by line quantization

In the index structure, each database vector is quantized to its nearest bin with a quantization error $\Delta_i$. To find the best vectors in the bin they need to be sorted based on their distance to the query vector. However, full $D$-dimensional distance calculations for each vector are too expensive. Similarly, re-ranking based on product quantized residuals [9] requires comparison to yet another codebook.

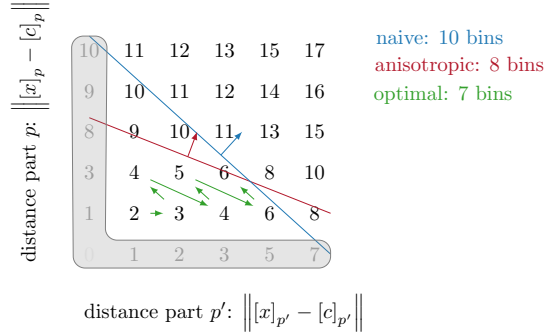

Figure 3: Merging the independent lookups from differents parts $p, p'$ to find the best bin-combination requires sorting all combinations. A Dijkstra-based traversal [4] (green) cannot be evaluated on a GPU due to its sequential nature, though it is the optimal sequence. Possible parallel approximations are a naive (blue) or a anisotropic (red) heuristic.

Inspired by the Johnson-Lindenstrauss lemma [8], we propose line quantization, where some of the information gathered during traversal is reused.

**Offline computation** Each vector (•) is quantized to the nearest projection (•) onto any line (■-■) through the level 1 centroids for each part, see Figure 4. For multiple parts, this quantization effectively spans hyper-planes. The distance of the query point to the line quantized vector can be evaluated exactly and efficiently using only one 2D triangle calculation per part.

In order to disambiguate the database vectors $x$ in these bins, we propose an approximation by projecting each vector part $[x]_p$ in the linear subspace $\text{span}([c_i]_p, [c_j]_p), c_i, c_j \in \mathcal{C}$ defined by the first level centroids $[c_i]_p$ and $[c_j]_p$ illustrated by ■ in Figure 4. We chose $[c_i]_p, [c_j]_p$ such that the quantization error $\delta_p$ is minimized. Therefore, when approximating each database vector part $[x]_p$ by $(1 - \lambda_p)[c_i]_p + \lambda_p[c_j]_p$ calculating the distance $d(y, x)$ does not rely on the values of $x$ but uses existing information from the tree-structure.

Using this approach, we store $\lambda_p$ and $(i_p, j_p)$ for each database vector using 1+1 bytes in our implementation. Storing the information of $\lambda_p$ and the indicies from $[c_i]_p, [c_j]_p$ in Figure 5 describes the approximation (•) of a vector (•). In fact, all information about database vector $x_i \in \mathcal{X}$ we need for the complete algorithm is encoded in the $3 \cdot P$ tuple

$$x_i \leftrightarrow (\lambda_1, \ldots, \lambda_P, i_1, \ldots, i_P, j_1, \ldots, j_P), \tag{7}$$

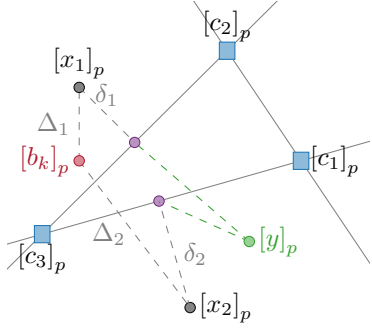

Figure 4: Line Quantization. In traditional PQ, each database vector $x_i$ (•) is projected onto the bin centroid (•) yielding an approximation error $\Delta_i$. Vectors in a bin would be indistinguishable wrt. a query $y$. We project $x_i$ onto (•) on the nearest line ■-■ which gives an approximation error $\delta_i$.

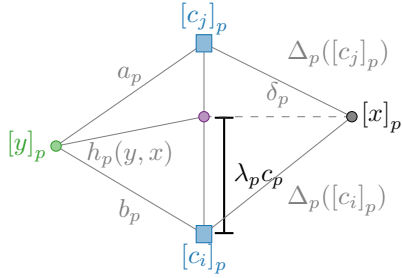

Figure 5: Exact query to line calculation. The database vector part ($[x]_p$, •) is projected onto (•) at the line ($[c_i]_p$, $[c_j]_p$) with error $\delta_p$. When re-ranking the exact distance $h_p$ between the query ($[y]_p$ •) and the quantized database point is obtained using triangulation. All necessary values are known as they are computed during tree traversal ($a_p, b_p$) or during database construction ($c_p, \lambda_p, (i, j)$).

which can be heavily compressed to about 2 bytes per part. While this scheme does not have the same compression rate as previous methods, it is the first to allow an efficient parallel re-ranking on the GPU by look-up from already computed values without any computational overhead.

We only need to store one small global lookup table of $P \times k_1 \times k_1$ precomputed distances between all pairs of level 1 centroids, i.e. $\|[c_s]_p - [c_t]_p\|_2^2$ for all $p, s, t$. This can be computed in the offline phase as it is *independent* of query vectors.

**Online computation** During tree traversal all distances between a query point $y \in \mathbb{R}^D$ and all level 1 centroids have already been computed as list of pairs $(i, d)_p^1$. The approximate distance to the database vector $x$ is computed given the triple $(\lambda_p, i_p, j_p)$, looking up $a_p$ and $b_p$ in the query's list, and $c_p$. The distance between $y$ and $x$ is approximated by

$$d(y,x)^2 = \sum_{p=1}^{P} d([y]_p, [x]_p)^2 \approx \sum_{p=1}^{P} h_p(y,x)^2 \quad (8)$$

$$\approx \sum_{p=1}^{P} \left(b_p^2 + \lambda_p^2 \cdot c_p^2 + \lambda_p \cdot (a_p^2 - b_p^2 - c_p^2)\right). \quad (9)$$

Note, that it is possible to compute the distance between a query and database vector by triangulation *exactly* up to the projection error $\delta_i$ as illustrated in Figure 5.

In practice we use different numbers of parts for the tree ($P_{\text{tree}} = 2$ or $4$) and for the line quantization ($P_{\text{line}} = 8, 16$ or $32$) for sufficiently precise re-ranking. Using exactly the same level 1 codebook with $p$ parts, we split each centroid part to get $p' = k \cdot p$ parts and compute the distances by aggregating the components.

## 4. GPU Implementation

Our approach is well suited to take advantage of GPU parallelism, which we implemented in CUDA. There are two levels of granularity of parallelism. The first is by processing multiple vectors in parallel, each vector with one block of threads. The second is by processing vector elements in parallel for all threads within the block.

Database bins are represented by a long, sorted array containing all vector IDs and a pointer array indicating where the vectors of each bin start. The pointer array is assembled by first computing a histogram of vectors over all bins and then computing the prefix sum [17]. In order to deal with a possibly excessive number of bins, we hash the bins using a simple modulus. As many bins contain zero vectors collision is simply ignored.

One kernel computes the distances to all level 1 centroids and sorts them using bitonic sort in shared memory. The second kernel does the same for the selected level 2 clusters based on the previous output. These two kernels are used both for sorting vectors into the bins as well as for kNN queries. For database vectors, a special kernel computes the optimal line projection (Sec. 3.3).

For each query vector, we then generate an ordered list of bins following the heuristic of Sec. 3.2. Each thread in a block computes and stores one bin ID. All empty bins are removed. Then, the kernel produces a list of potential vector IDs, each thread is responsible for copying all vectors IDs of one bin. The final kernel calculates the distances for the re-ranking using the line quantization and outputs the re-ranked list of IDs. Here, re-ranking one vector is executed by one warp each.

| method | ms | R@1 | R@10 | R@100 | su |
|---|---|---|---|---|---|
| FLANN [16] | 5.32 | 0.97 | - | - | ×9.6 |
| LOPQ [11] | 51.1 | 0.51 | 0.93 | 0.97 | ×1 |
| IVFADC* | 11.2 | 0.28 | 0.70 | 0.93 | ×4.5 |
| $PQT_1$ (CPU) | 4.89 | 0.45 | 0.86 | 0.98 | ×10.4 |
| $PQT_2$ (CPU) | 5.74 | 0.98 | (exact re-ranking) | | ×8.9 |
| PQT (GPU) | 0.02 | 0.51 | 0.83 | 0.86 | ×2555 |
| GPU brutef. | 23.7 | 1 | 1 | 1 | ×2 |

Table 1: Performance on the SIFT1M dataset using different methods. Reported query times include query + re-ranking times. The GPU implementation uses the first $2^{12}$ vectors from the proposed bins and $(64 \cdot 8)^4$ bins. The reported CPU performance is base on $(8 \cdot 4)^2$ bins. Speedup (su) is reported relative to the slowest method. $PQT_2$ is $PQT_1$ but with additional exact re-ranking. (*) indicates that the timing was reported by the authors. R@$n$ means, the correct vector is within the first $n$ returned vectors from the algorithm.

## 5. Results

We now present the results of the PQT evaluated on several standard benchmark sets. All reported CPU query times were obtained from a single-thread C++ implementation using SSE2 instructions. Results of our GPU implementations are obtained with a NVIDIA GTX Titan X.

We use the publicly available benchmark SIFT1M, SIFT1B datasets [10], of $10^6, 10^9$ 128-dim vectors and GIST1M [9] of $10^6$ 960-dim vectors. For the codebook training process we used the first 100K/1M vectors from the respective datasets. It was not possible to obtain any results on GIST1M using FLANN in Table 2.

### 5.1. Query times and Recall

We compared our implementation with the available implementations of [11] and [4]. Due to the approximation nature of these algorithms and discrete parameter space it is not trivial to find parameter settings which produce the same accuracy for timing comparisons. Therefore, we choose a highly optimized GPU-based exhaustive search as a strong baseline method. The accuracy is measured in *recall* R@$x$, which is the fraction of nearest neighbors found in the first $x$ proposed vectors after re-ranking.

Table 1 gives the average query time in milliseconds obtained on the same machine using public available code. Compared to all PQ-based approaches [4, 11] our approach ($P_{\text{line}} = 32$) is faster on the CPU at similar accuracy. Allowing [11] to use more memory consumption for re-ranking slows down the query process. Note that the reported time of [11] excludes all intensive operations like the multiplication of query vector with a $D \times D$ rotation matrix, which was pre-computed.

| method | avg. (ms) | R@100 |
|---|---|---|
| SH [5] | 22.7 | 0.132 |
| IVFADC | 65.9 | 0.744 |
| FLANN | not possible | |
| PQT(CPU) | 63 | 0.83 |

Table 2: Performance of the GIST1M dataset using different methods. PQT uses 128 parts for re-ranking.

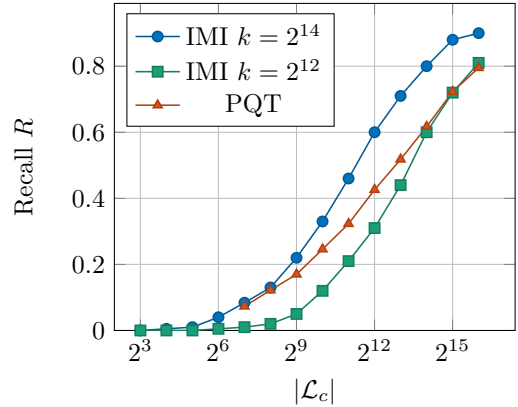

Figure 6: Recall rates on the SIFT1B data set ($p = 4, k_1 = 32, k_2 = 16, w = 8$) with ordering of bins. The recall from PQT is *without* a reranking step. Even with significant lower query time, our approach is comparable in quality to the inverted multi-index with $k = 2^{12}$.

On the GPU, sorting the SIFT1M vectors into the bins takes 1051ms, performing the line quantization for these 1M vectors about 458ms ($p = 4, k_1 = 16, k_2 = 8, w = 8$). The processing time for one query is roughly 39 microseconds, split into 4% traversal, 35% bin selection, 11% vector proposal, and 50% re-ranking. In our implementation the maximum number of sortable vectors on the GPU per query is currently limited to 4096 during re-ranking. Applying different algorithms, this restriction could be removed.

With the right configuration of bins, high recall values can even be achieved on the SIFT1B data set (Figure 6). Because the full data set did not fit on the GPU, the data base was build in waves of 1M vectors, aggregating the information on the CPU. With file I/O this took about 144min. On a NVIDIA GTX Titan X with 12GB of RAM one can upload the resulting DB structure, i.e. bin sizes and vector IDs per bin. For the SIFT1B dataset it was essential to re-sort the proposed bins by the actual distance. This slowed down query time to 0.027ms in total without re-ranking. The recall rate of our approach is R@10=0.35. For the re-ranking we directly accessed the CPU main memory from the GPU

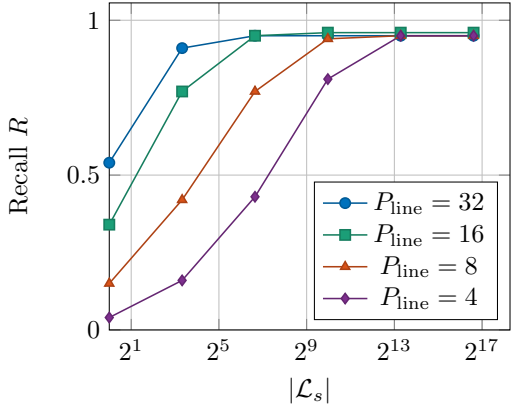

Figure 7: Line Quantization. The different curves show the recall of the SIFT1M dataset for varying values of $P_{\text{line}}$ using $p=2, k_1=16, k_2=8, w=4$. A query took $3,4,6,9$ms on a single CPU with $|\mathcal{L}_c| < 20000$.

| $P_{\text{line}}$ | min distor. | max distor. | avg. distor. | time (ms) |
|---|---|---|---|---|
| 2 | 10874.9 | 179870 | 30534.6 | 2.0 |
| 4 | 8967.8 | 166722 | 26257.9 | 2.6 |
| 8 | 6709.2 | 145082 | 19719.4 | 3.6 |
| 16 | 3318.3 | 84640 | 10509.7 | 5.3 |
| 32 | 1035.3 | 39143 | 3686.71 | 8.5 |

Table 3: Squared Line-quantization error (distortion $\delta$) by projecting each $x \in \mathcal{X}$ onto a line using $p=2, k_1=16, k_2=8, h_1=4$ for the SIFT1M data set. Last column gives the average query time.

resulting in a total query time of 0.15ms.

Memory is the limiting factor for the maximum number of actual bins. We apply hashing to 100M bins. Increasing the number of parts $P$ or introducing a further level into the tree would further boost the number of bins – at the same time, also the number of bins to be visited in the vicinity would drastically increase and slow down the system.

### 5.2. Precision of Line Quantization

We tested the performance of encoding each database vector $x \in \mathcal{X}$ by its projection onto a line for different numbers of parts used for line quantization (see Figure 7 and Table 3). The recall rate increases with the number of line parts, $P_{\text{line}}$. Low quantization errors with moderate computational and storage effort are obtained with $P_{\text{line}} = 16$. Note, that the necessary data for each query vector is directly assembled during the tree traversal without the need for any further quantization computation. See the supplementary material for re-ranking results on MNIST (784 dimensions).

## 6. Conclusion

In conclusion, we introduce a new method for efficient similarity search on large, high dimensional datasets. We propose a two level Product Quantization Tree for quickly indexing large numbers of bins with minimal memory and computation overhead. We combined this with a novel re-ranking method based on closest-line projections, and a bin ordering heuristic. The tree structure provides all intermediate values, which accelerate the re-ranking procedure.

Our prototype implementation demonstrates improvement in accuracy and speed over state-of-the art methods for ANN queries. We demonstrated the scalability from competitive performance to FLANN [16] in small benchmark sets (SIFT1M) and outperform to the state-of-the-art methods at the challenging BigANN benchmark set containing one billion vectors of dimension 128, as well as in datasets with high dimensionality (GIST1M). By construction, the proposed approach can easily be implemented as well on the GPU, which evaluates to a significant speedup against previous methods.

While our method worked well in the examples and datasets we tried, there are many avenues for future research. For example, it is possible that other tree structures featuring different combinations of PQ and VQ, or even these methods in combination with different approaches such as KD-trees, or LSH would be an interesting area of future research. Additionally, performing efficient on-the-fly updates to the database vectors, and resulting tree structure would be another area for future work.

**Acknowledgement** This work has been partially supported by the DFG Emmy Noether fellowship Le 1341/1-1 and an NVIDIA hardware grant.